\documentclass[final]{cvpr}

\usepackage{times}
\usepackage{epsfig}
\usepackage{graphicx}
\usepackage{amsmath}
\usepackage{amssymb}
\usepackage{caption}
\usepackage{bbm}
\usepackage{mathtools}

\usepackage[pagebackref=true,breaklinks=true,colorlinks,bookmarks=false]{hyperref}

\def\cvprPaperID{5522} 
\def\confYear{CVPR 2021}

\usepackage[dvipsnames]{xcolor}

\begin{document}

\title{Pixel-aligned Volumetric Avatars}

\author{Amit Raj$^1$ \quad
Michael Zollh\"{o}fer $^2$ \quad
Tomas Simon $^2$ \quad
Jason Saragih $^2$ \quad \\
Shunsuke Saito $^2$ \quad
James Hays $^1$ \quad
Stephen Lombardi $^2$ \quad
\\ \quad \\
$^1$ Georgia Institute of Technology \quad
$^2$ Facebook Reality Labs \quad
}
\twocolumn[{%
\renewcommand\twocolumn[1][]{#1}%
\vspace{-2em}
\maketitle
\vspace{-1em}
\begin{center}
    \centering
    \includegraphics[width=0.90\linewidth]{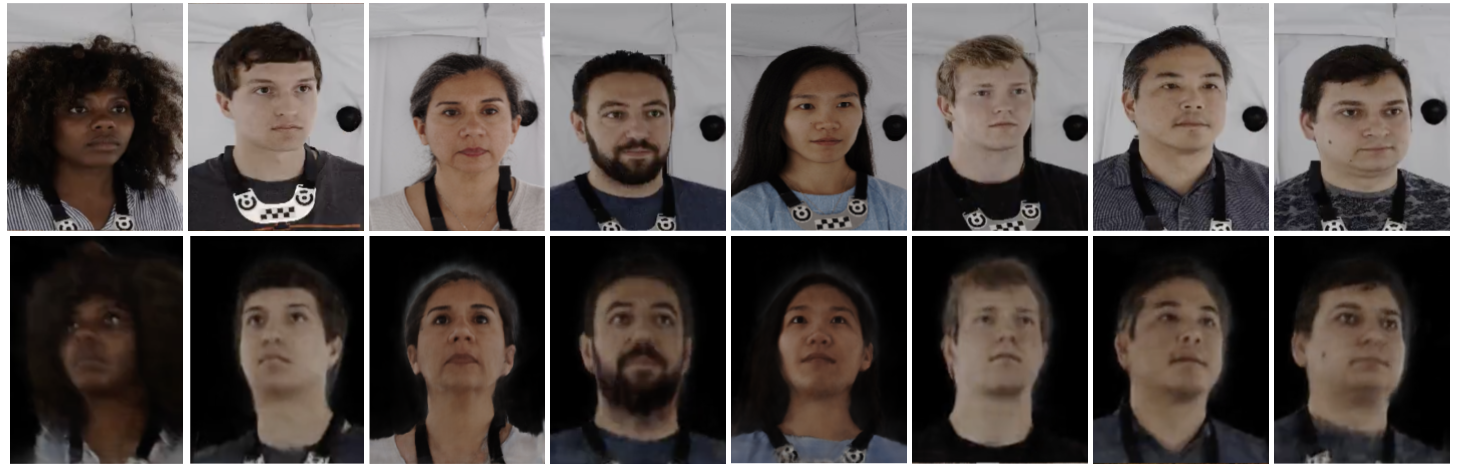}
    \captionof{figure}
    {
    We present a novel approach for the prediction of volumetric avatars of human heads from a small number of example views.
    Our model enables view synthesis for unseen identities and is able to generate faithful facial expressions.
    }
    \label{fig:teaser}
\end{center}%
}]

\begin{abstract}
%
Acquisition and rendering of photo-realistic human heads is a highly challenging research problem of particular importance for virtual telepresence.
Currently, the highest quality is achieved by \emph{volumetric} approaches trained in a person-specific manner on multi-view data.
These models better represent fine structure, such as hair, compared to simpler mesh-based models.
Volumetric models typically employ a global code to represent facial expressions, such that they can be driven by a small set of animation parameters.
While such architectures achieve impressive rendering quality, they can not easily be extended to the multi-identity setting.
In this paper, we devise a novel approach for predicting volumetric avatars of the human head given just a small number of inputs.
We enable generalization across identities by a novel parameterization that combines neural radiance fields with local, pixel-aligned features extracted directly from the inputs, thus side-stepping the need for very deep or complex networks.
Our approach is trained in an end-to-end manner solely based on a photometric re-rendering loss without requiring explicit 3D supervision.
We demonstrate that our approach outperforms the existing state of the art in terms of quality and is able to generate faithful facial expressions in a multi-identity setting.
\end{abstract}

\section{Introduction}
The acquisition and rendering of photo-realistic human heads is a highly challenging research problem with high significance for virtual telepresence applications.
Human heads are challenging to model and render due to their complex geometry and appearance properties: sub-surface scattering of skin, fine-scale surface detail, thin-structured hair, and the human eyes as well as the teeth are both specular and translucent.
Most existing approaches require complex and expensive multi-view capture rigs (with up to hundreds of cameras) to reconstruct even a person-specific model of a human head.

Currently, the highest-quality approaches are those that employ \emph{volumetric} models rather than a textured mesh, since they can better learn to represent fine structures on the face like hair, which is critical to achieving a photo-realistic appearance.
These volumetric models \cite{Lombardi2019} typically employ a global code to represent facial expressions or only work for static scenes \cite{Mildenhall2020,Liu2020}.
While such architectures achieve impressive rendering quality, they can not easily be adapted to a multi-identity setting.
A global code, as is used to control expression, is not sufficient for modeling identity variation across subjects.
There has been significant progress of late in using implicit models to represent scenes and objects.
These models have the advantage that the scene is represented as a parametric function in a continuous space, which allows for fine-grained inference of geometry and texture \cite{Saito2019}.
But these approaches can not model view-dependent effects and it is challenging to represent for example hair with a textured surface.
The approach of Sitzmann et al.~\cite{Sitzmann2019b} can generalize across objects, but only at low resolutions and can only handle purely Lambertian surfaces, which is not sufficient for human heads.
Despite the recent success and advantages of such scene representation approaches, there are several limitations.
In particular, most of the above methods train a network to model only a single scene or object.
Methods which can generate multiple objects are typically limited in terms of quality and resolution of the predicted texture and geometry. 

We present pixel-aligned volumetric avatars (PVA), a novel framework for the estimation of a volumetric 3D avatar from only a few input images of a human head.
Our approach is able to generalize to unseen identities at test time.
Methods such as Scene Representation Networks (SRNs) \cite{Sitzmann2019a}, which generate a set of weights from a global image encoding (i.e., a single latent code vector per image), have difficulty generalizing to local changes (e.g., facial expressions) and fail to recover high-frequency details even when these are visible in the input images.
This is because the global latent code summarizes information in the image and must discard some information to generate a compact encoding of the data.
To improve generalization across identities, we instead parameterize the volumetric model via local, pixel-aligned features extracted from the input images.

We show that our model can synthesize novel views for unseen identities and expressions while preserving high frequency details in the rendered avatar. 
To summarize, our contribution are:
\begin{itemize}
\item We introduce a novel pixel-aligned radiance field that predicts implicit shape and appearance from a sparse set of posed images.
\item Our model generalizes to
unseen identities and expressions at test time.
\item We demonstrate state of the art performance on novel view synthesis compared to recent approaches.
\end{itemize}


\section{Related Work}
Generating avatars from images has a long history in computer vision and graphics.
Traditional methods employ mesh-based representations and physics-inspired models of how faces deform and interact with light, while more recent approaches employ deep learning to overcome some of the limitations of classical techniques.
We discuss several classes of methods below and compare them to ours.
\begin{figure*}
    \centering
    \includegraphics[width=1.0\linewidth]{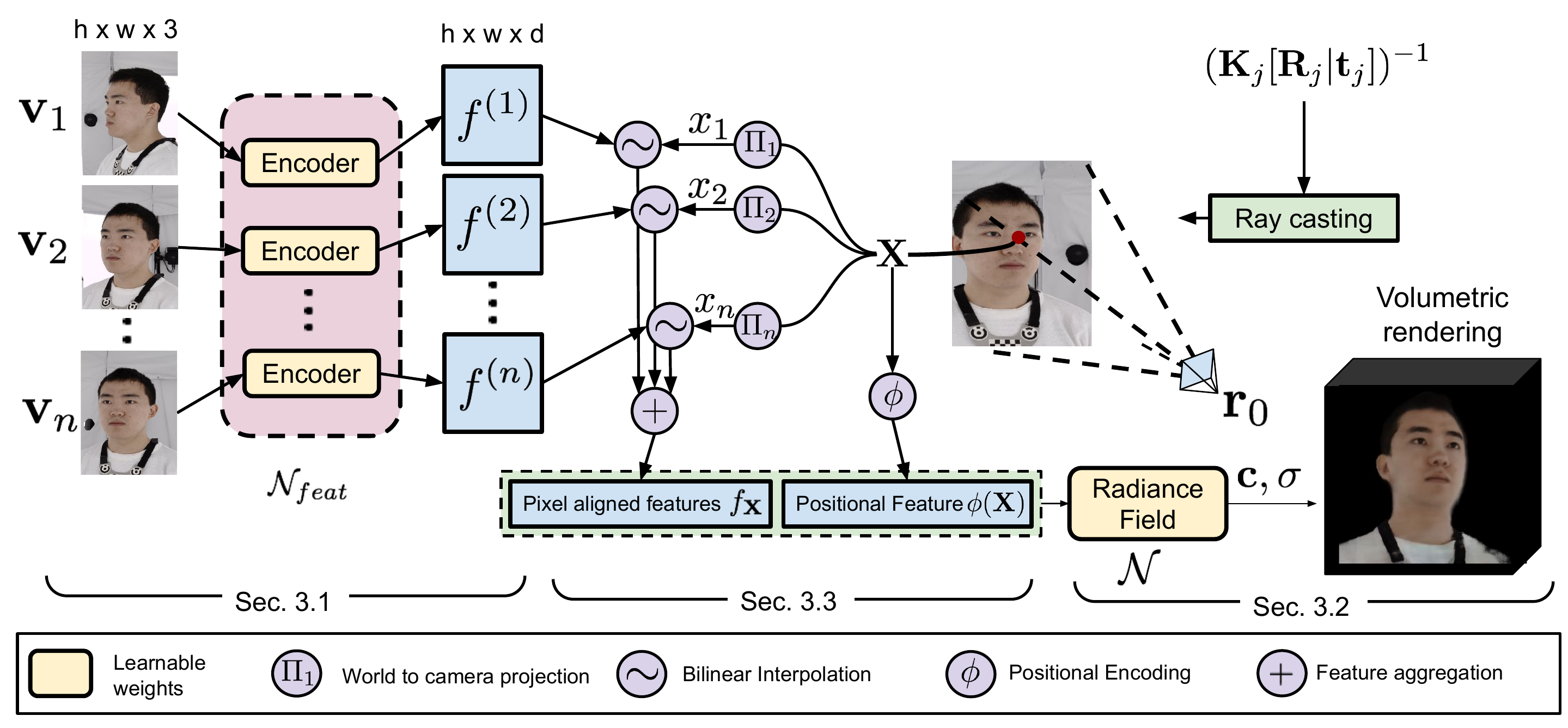}
    \caption{Overview of the proposed approach. Given a target viewpoint and a set of conditioning images, our novel approach employs local, pixel-aligned features that are extracted from the inputs to condition a multi-identity neural radiance field. Volume rendering is employed to generate an image of the subject from the target viewpoint.}
    \label{fig:framework}
\end{figure*}
\paragraph{Mesh-based Approaches}
%
Active Appearance Models (AAMs) was among the first face models capable of modeling facial expressions, although it was originally used as a statistical joint shape and appearance model for human faces \cite{Cootes1998}, and later extended to 3D faces \cite{Blanz1999}.
%
Deep Appearance Models \cite{Lombardi2018,nagano2018pagan} create a 3D morphable model using deep networks to create an extremely high-quality and driveable face model.
However mesh-based methods struggle with rendering thin structures like hair, which are critical for realistic human face rendering.
%
Mesh-based methods have been extended in a number of ways to improve quality and expressiveness, though they typically share similar disadvantages.
Notably, mesh-based models require a fixed topology, which poses problems for modeling hair, which can vary dramatically from one person to another.
Furthermore, mesh-based methods have hard triangle boundaries which can look unpleasant for soft features.
Finally, optimizing meshes to match the appearance of arbitrary shapes is still a difficult problem.
Efforts in differentiable rasterizers \cite{chen2019interpolation, liu2019soft, kato2018neural, kanazawa2018learning} have shown impressive results in generating meshes from single and multi-view images without 3D supervision, but the generated meshes usually have restrictions in terms of topology and fail to capture high frequency details.
Furthermore, they are limited in terms of the textures that can be represented. 
In contrast, our method is able to capture arbitrary topology (as seen in expressions and hairstyles) and captures high frequency texture details better, since it is able to use pixel-level information more efficiently.

\paragraph{Image-based Methods}
Recently, there has been a great deal of progress in high-quality controllable face synthesis \cite{Karras2019,viazovetskyi2020stylegan2, abdal2019image2stylegan}.
However, these image-based methods work with mostly frontal faces and have difficulty explicitly controlling the viewpoint and expression of the synthesized images.
Without giving the network a notion of 3D space, it is difficult for the methods to generalize without many training images.
StyleRig \cite{tewari2020stylerig} enables parameteric control of StyleGAN generated imagery.
However, the results are not multi-view consistent and the approach does not work on real images.

\paragraph{Voxel-based Methods}
Methods such as \cite{zhu2018visual,nguyen2018rendernet,sitzmann2019deepvoxels} learn an intermediate 3D voxel grid of features and a 3D-2D projection operation to synthesize images.
Transformable bottleneck networks \cite{olszewski2019transformable} present a method that learns a bottleneck of 3D features that can be manipulated directly to enable a variety of applications.
However, the primary problem with these voxel-based approaches are their inability to scale to higher resolution due to memory restrictions.
We eschew the problem of capacity by learning a multi-layer perceptron (MLP) that directly translates 3D locations and pixel-aligned features to color and occupancy.

\paragraph{Implicit Methods}

Works such as \cite{yariv2020multiview} use explicit 3D information during training. 
PIFu/PIFuHD \cite{Saito2019,Saito2020} models human bodies with an implicit function evaluated at the depth of a point.
It is capable of rendering human bodies with high quality.
A key insight is to use pixel-aligned features to retain high-frequency detail.
We leverage this insight, but our method does not require 3D supervision.
Scene Representation Networks (SRNs) \cite{Sitzmann2019b} model scenes with a learned SDF and do not require 3D supervision.
We do not assume a well-defined surface through an SDF but rather a semi-transparent representation that can better model hair and thin structures.
The authors of \cite{nguyen2019hologan, henzler2019escaping} learn an implicit representation of geometry from natural images in an unsupervised manner to allow novel view synthesis.
These methods are limited in the degree of multi-view consistency that can be achieved. 
TextureFields \cite{OechsleICCV2019} learn to transfer textures from an exemplar image to a source mesh to allow novel view synthesis.
We eschew the need for a mesh at inference time by learning an implicit representation of geometry.

\paragraph{Neural Rendering}
Many neural rendering models have been proposed recently that better represent thin structures, like hair and clothes.
Neural volumes \cite{Lombardi2019} and NeRF \cite{Mildenhall2020} are two recently introduced methods that model objects with a semitransparent volume and have shown the ability to model thin structures well.
Neural Volumes can also model dynamic scenes.
NeRF-W \cite{martin2020nerf} extends the work of \cite{Mildenhall2020} to a conditional setting to models scenes under different lighting with same underlying geometry.
However, these methods fail to generalize to novel identities.
Inspired by insights from NeRF and PIFu, we demonstrate a framework that handles multiple identities by relying on pixel-aligned features.
GRAF \cite{schwarz2020graf} learns a conditional radiance field in an unsupervised manner by disentangling a global shape and appearance code which limits its ability to model local shape and texture deformations.
Other works focus on speeding up NeRF using a sparse Octree structure \cite{liu2020neural}.
%
%
We refer the readers to the recent STAR of Tewari et al.~\cite{tewari2020state} for an in-depth treatment of recent neural rendering methods.

\section{Approach}


We present a Pixel-aligned Volumetric avatars(PVA). An  implicit model of faces that is learned from a multi-view image collection, see Fig.~\ref{fig:framework}. Our model can generate novel views of unseen identities from one or more example images. The framework consists of two main components. The first is a shallow convolutional encoder-decoder ($\mathcal{N}_{feat})$ network that takes as input one or more images ($\mathbf{v}_i$) of a person from a known viewpoint $\{\mathbf{K}_i, \left[\mathbf{R|t}\right]_i\}$ and produces pixel-aligned feature maps $f^{(i)}$. The second component is a radiance field network ($\mathcal{N}$) that converts 3D location and pixel-aligned features to color and opacity. To render the radiance field, we march along the camera ray of each pixel in the target view $j$, defined by $\{\mathbf{K}_j, \left[\mathbf{R|t}\right]_j\}$, accumulating the color and occupancy produced by $\mathcal{N}$ at each point.
We train our approach based on a multi-identity training corpus using gradient descent.
To this end, we minimize the $L_2$ loss between predicted images and the corresponding ground truth.

\subsection{Pixel-aligned Radiance Fields}
We employ a pixel-aligned scene representation modeled as a neural network.
Concretely, for a conditioning view $\mathbf{v}_i \in \mathbbm{R}^{h \times w \times 3}$ we define functions 
\begin{align}
 f^{(i)} &= \mathcal{N}_{\mathrm{feat}}(\mathbf{v}_i) \\
 \left\{ \mathbf{c}, \mathbf{\sigma} \right\} &= \mathcal{N}(\phi(\mathbf{X}),f_{\mathbf{X}}) 
\end{align}
where $\phi(\mathbf{X}) : \mathbbm{R}^3 \rightarrow \mathbbm{R}^{6\times l}$ is the positional encoding of $\mathbf{X} \in \mathbbm{R}^3$ as in \cite{Mildenhall2020} with $2 \times l$ different basis functions, $f^{(i)}\in \mathbbm{R}^{h \times w \times d}$ is the feature map of $\mathbf{v}_i$, $d$ the number of feature channels, $h$ and $w$ are image height and width, and
$f_\mathbf{X} \in \mathbbm{R}^{d'} $ is the aggregated image feature associated with the point $\mathbf{X}$ as explained in the next section.
For each feature map $f^{(i)}$, we obtain $f_\mathbf{X}^{(i)} \in \mathbbm{R}^d$ by projecting 3D point $\mathbf{X}$ along the ray using camera intrinsic and extrinsic parameters $\mathbf{K}, \mathbf{R},\mathbf{t}$ of that particular viewpoint,
\begin{align}
x_i &= \Pi(\mathbf{X};\mathbf{K}_i\left[\mathbf{R|t}\right]_i),\label{eq:pixelproj} \\
f_\mathbf{X}^{(i)} &= \mathcal{F}(f^{(i)} ; x_i)
\end{align}
where $\Pi$ is a perspective projection function to camera pixel coordinates, and $\mathcal{F}(f,x)$ is the bilinear interpolation of $f$ at pixel location $x$.

\subsection{Volume Rendering}
%
For each given training image  $\mathbf{v}_j$ with camera intrinsics $\mathbf{K}_j$  and rotation and translation $\mathbf{R}_j, \mathbf{t}_j$, the predicted color of a pixel $p \in \mathbb{R}^2$ for a given viewpoint in the focal plane of the camera and center $\mathbf{r}_0 \in \mathbb{R}^3$ is obtained by marching rays into the scene using the camera-to-world projection matrix,
$
\mathbf{P}^{-1} = [\mathbf{R}_i|\mathbf{t}_i]^{-1}\mathbf{K}_i^{-1} 
$ with the direction of the rays given by,
\begin{equation}
    \mathbf{d} = \frac{\mathbf{P}^{-1}p - \mathbf{r}_0 }{\lVert{\mathbf{P}^{-1}p - \mathbf{r}_0}\rVert}.
\end{equation}
Note that in order to help the network focus its capacity on modeling the content of the scene, all camera extrinsics are relative to the computed head pose, which is found via traditional head registration.

We then accumulate the radiance and opacity along the ray $\mathbf{r}(t) =\mathbf{r}_0 + t\mathbf{d}$ for $t \in [t_{\mathrm{near}},t_{\mathrm{far}}]$ as defined in NeRF~\cite{Mildenhall2019} as follows:
\begin{equation}
    \mathbf{I}_{rgb}(p) = \int_{t_{\mathrm{near}}}^{t_{\mathrm{far}}} \mathbf{T}(t)\sigma(\mathbf{{r}}(t))\mathbf{c(r}(t),\mathbf{d}) dt
\end{equation}
where,
\begin{align}
    \mathbf{T}(t) &= \operatorname{exp}\left(-\int_{t_{\mathrm{near}}}^{t} \mathbf{\sigma(r}(s))ds\right)
\end{align}

In practice we uniformly sample a set of $n_s$ points $t\sim[t_{near}, t_{far}]$. We set $\mathbf{X}=\mathbf{r}(t)$ and use the quadrature rule to approximate the integral.
We also define $\mathbf{I}_{\alpha}(p)$ as,
\begin{equation}
    \mathbf{I}_{\alpha}(p) = \sum_{i=1}^{n_s} \alpha_i \prod_{j=1}^i(1-\alpha_j)
\end{equation}
where $\alpha_i = 1-\operatorname{exp}(-\delta_i\sigma_i)$ with $\delta_i$ being the distance between the $i+1$-th and $i$-th sample point along the ray.


\subsection{Multi-view Feature Aggregation}

A critical component of our method is how to fuse pixel-aligned features $f^{(i)}_\mathbf{X}$ from multiple images to help the network best use this information.

\subsubsection{Fixed number of conditioning views}

In a multi-view setting with known camera viewpoints and a fixed number of conditioning views we can aggregate the features by simple concatenation \cite{Lombardi2019}. Concretely, for $n$ conditioning images $\{\mathbf{v}_i\}_{i=1}^n$ with corresponding rotation and translation matrices given by $\{\mathbf{R}_i\}_{i=1}^n$ and $\{\mathbf{t}_i\}_{i=1}^n$. We obtain $n$ features $\{f_{\mathbf{X}}^{(i)}\}_{i=1}^n$ for each point $\mathbf{X}$ as in Eq.~\ref{eq:pixelproj} and generate the final feature as follows,
\[
f_\mathbf{X} = [f_\mathbf{X}^{(1)} \bigoplus f_\mathbf{X}^{(2)} ... \bigoplus f_\mathbf{X}^{(n)}] 
\]
where $\bigoplus$ represents concatenation along the depth dimension.
This preserves feature information from all the viewpoints, leaving the MLP to figure out how to best combine and employ the conditioning information.

\subsubsection{Variable number of conditioning views}

The more interesting use case is to make the model agnostic to viewpoint and number of conditioning views. Simple concatenation as above is insufficient in this case, since we do not know the number of conditioning views a priori, leading to different feature dimensions during inference time. To summarize features for a multi-view setting we need a permutation invariant function $\mathcal{G} : \mathcal{R}^{n \times d} \rightarrow \mathcal{R}^d $ such that for any permutation $\psi$,
\[
\mathcal{G}([f^{(1)},f^{(2)},...,f^{(n)}]) = \mathcal{G}([f^{\psi(1)},f^{\psi(2)},...,f^{\psi(n)}]).
\]

A simple permutation invariant function for feature aggregation is the mean of the sampled features (as employed in PIFu~\cite{Saito2019}). This is a reasonable aggregation procedure when we have depth information during training. However, since we have inherent depth ambiguity (since the points are projected onto the feature image before sampling) we find that this kind of aggregation produces artifacts. Fig.~\ref{fig:camera-sum} shows an example of this behavior.

This simple mean of image features does not consider camera information, which may help the network use the conditioning information more effectively. To inject viewpoint information into the feature, we learn another network $\mathcal{N}_{cf} : \mathcal{R}^{d+7} \rightarrow R^{d'}$ that takes the feature vector and the camera information ($\mathbf{c}_i$) and produces a \emph{camera-summarized} feature vector. These modified vectors are then averaged for all conditioning views as follows
\begin{align}
    f_\mathbf{X}^{\prime(i)} &= \mathcal{N}_{cf}(f_\mathbf{X}^{(i)}, \mathbf{c}_i) \\
    f_\mathbf{X} &= \frac{1}{n}\sum_{i=1}^n  f_\mathbf{X}^{\prime(i)}
\end{align}
The advantage of this approach is that the camera-summarized features can take likely occlusions into account before the feature average is performed. The camera information is encoded as a 4D rotation quaternion and 3D camera position.

\subsection{Background Model}
\begin{figure}
    \centering
    \includegraphics[width=1.0\linewidth]{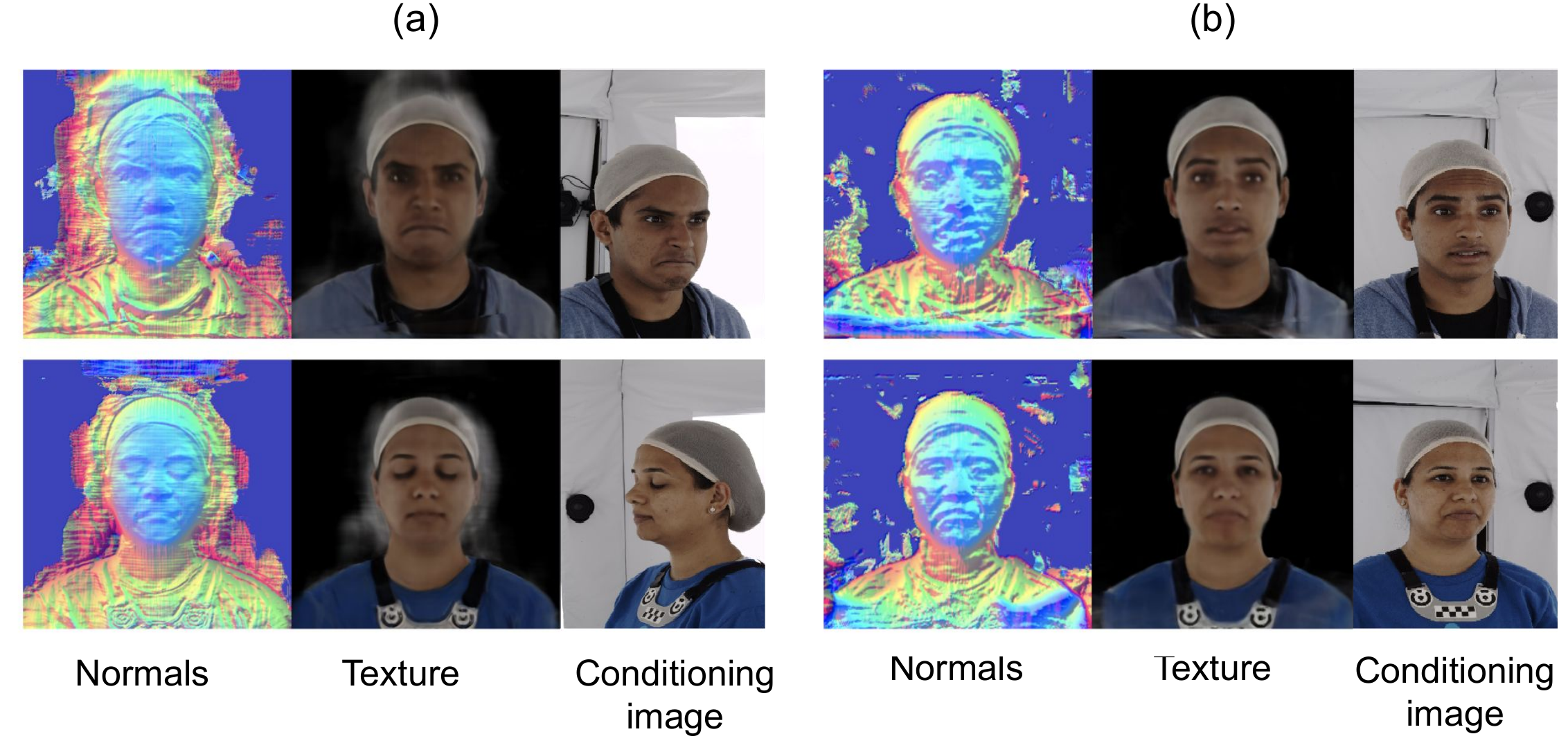}
    \vspace*{-0.2cm}
    \caption{Background ablation. (a) Without background estimation (b) Ours. Our learned background model leads to better reconstruction results.}
    \label{fig:bg-halo}
\end{figure}

To avoid learning parts of the background in the scene representation, we define a background estimation network: $\mathcal{N}_{bg}: \mathcal{R}^{n_c} : \rightarrow \mathcal{R}^{h \times w \times 3}$ to learn a per-camera fixed background. 
Particularly, we predict the final image pixels as
\begin{equation}
    \mathbf{I}_{p} = \mathbf{I}_{rgb}  + (1-\mathbf{I}_{\alpha})\mathbf{I}_{bg}
\end{equation}

with $\mathbf{I}_{bg} = \mathbf{\bar{I}}_{bg} + \mathcal{N}_{bg}(C_i)$ for camera $C_i$ where $\mathbf{\bar{I}}_{bg}$ is an initial estimate of the background extracted using inpainting. These inpainted backgrounds are often noisy leading to `halo' effects around the head of the person (Fig.~\ref{fig:bg-halo}). Our background estimation model learns the residual to the inpainted background. This has the advantage of not needing a high capacity network to account for the background.

\subsection{Color Correction Model}
The different camera sensors have a slightly different response to the same incident radiance despite the fact that they are the same camera model.
If nothing is done to address this, the intensity differences end up baked into the scene representation $\mathcal{N}$, which will cause the image to unnaturally brighten or darken from certain view points.
To address this, we learn a per-camera bias and gain value.
This allows the system to have an `easier' way to explain this variation in the data.

\subsection{Loss Function}
For ground truth target images $\mathbf{v}_j$, we train both the radiance field and feature extraction network using a simple photo-metric reconstruction loss:
\[
\mathcal{L}_{\mathrm{photo}} = \lVert\mathbf{I}_{p_j} - \mathbf{v}_j \rVert_2 ~.
\]
Note, our approach is trained in an end-to-end manner solely based on this 2D re-rendering loss without requiring explicit 3D supervision.




\begin{figure*}
    \centering
    \includegraphics[width=0.32\linewidth]{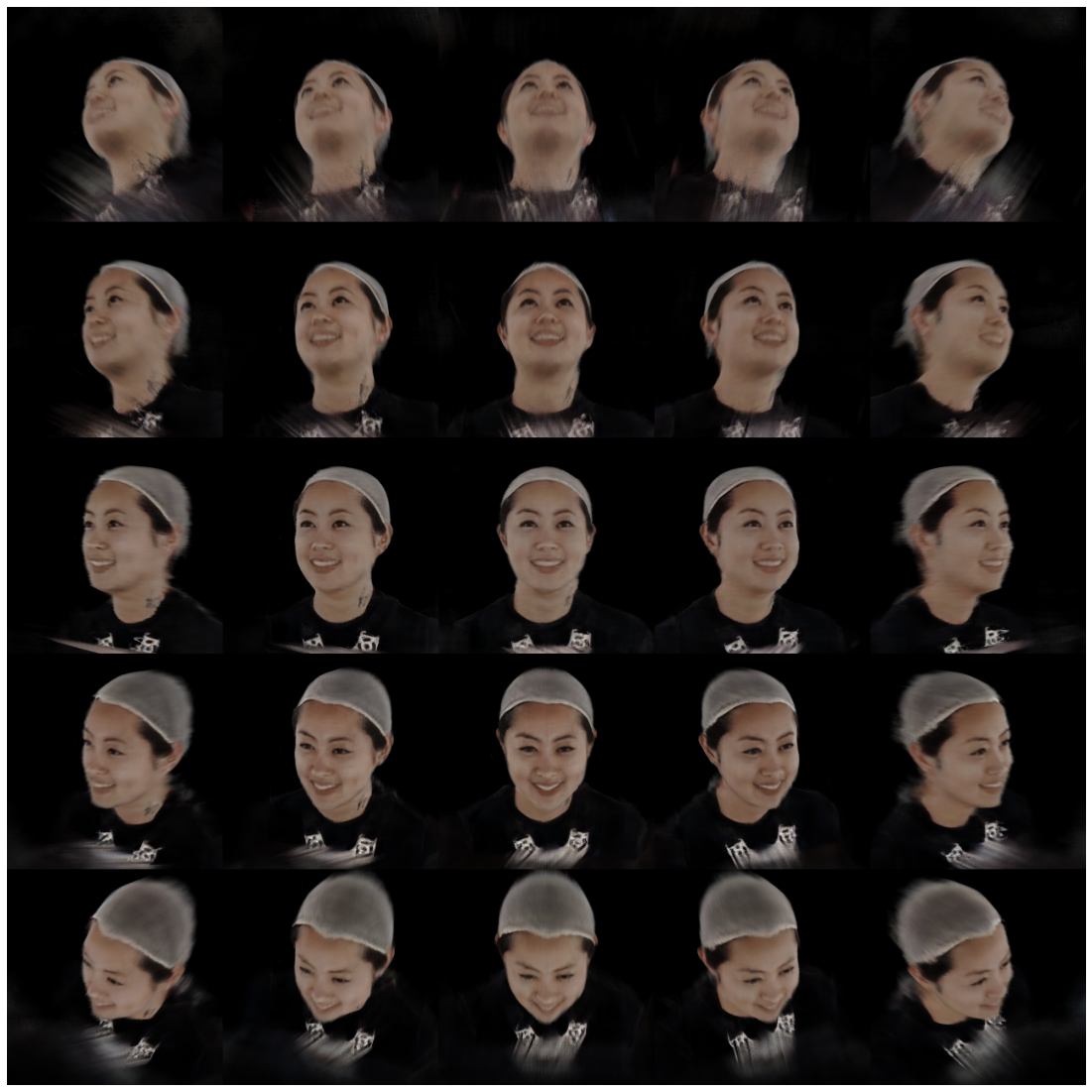}
    \includegraphics[width=0.32\linewidth]{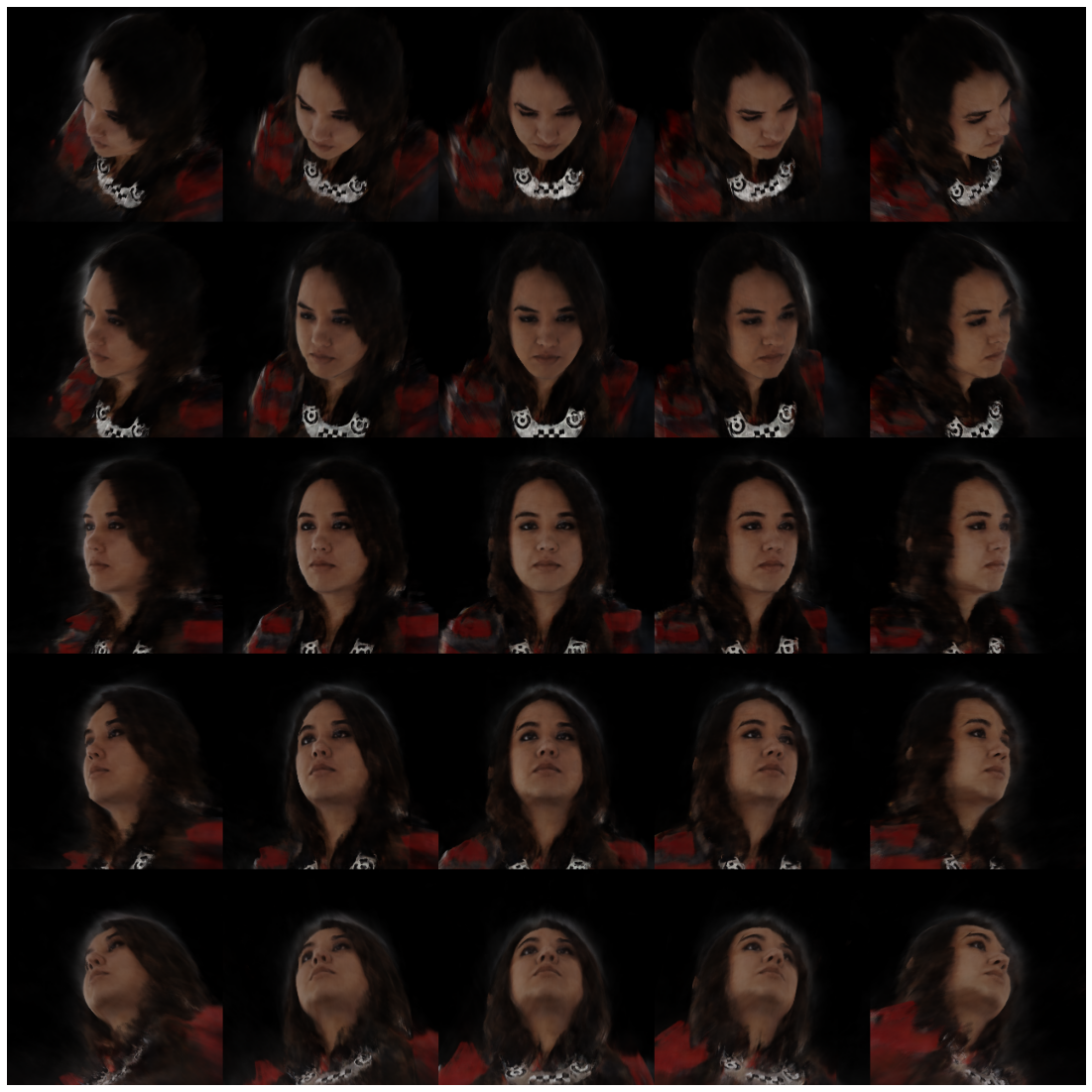}
    \includegraphics[width=0.32\linewidth]{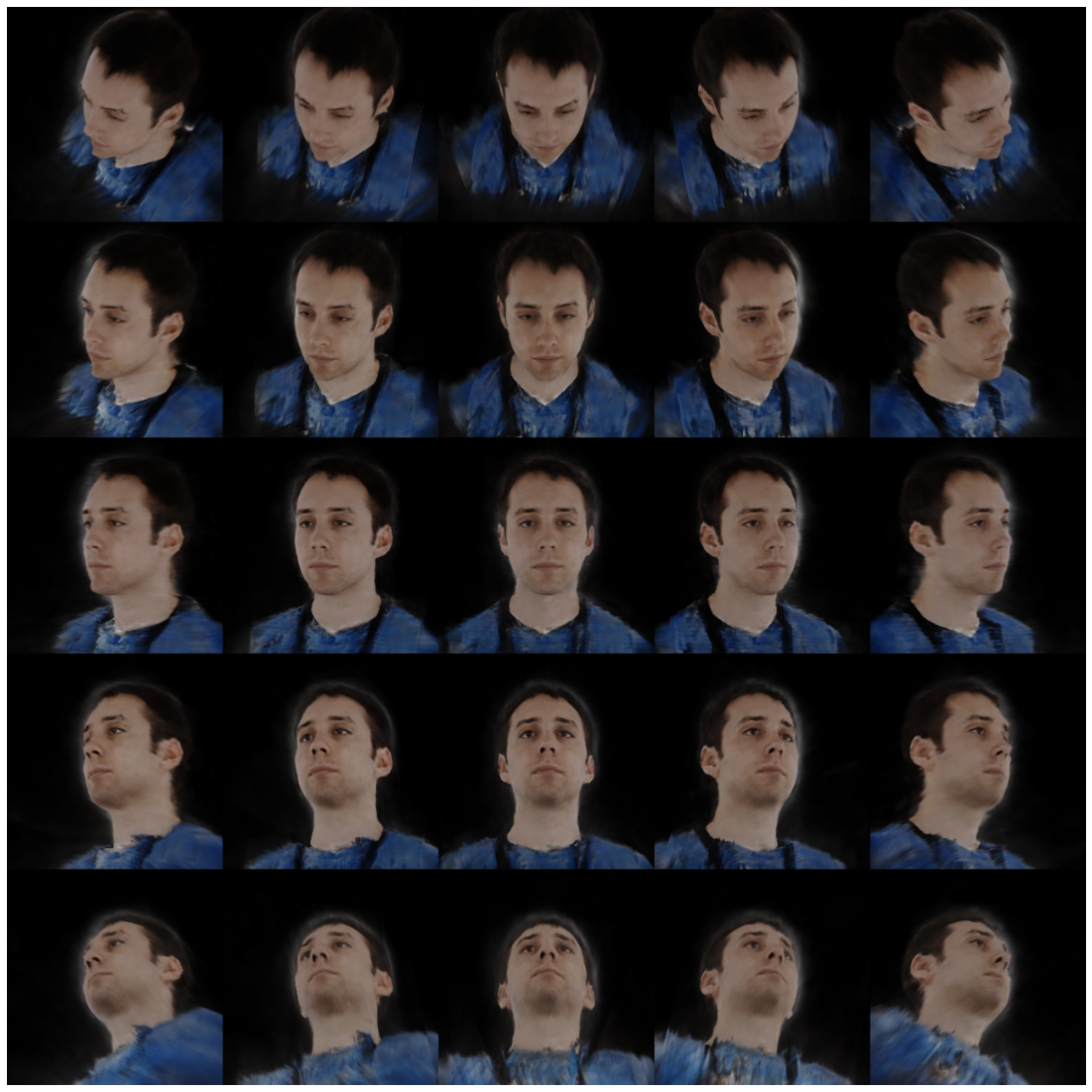}
    \caption{We demonstrate novel view synthesis of unseen identities using pixel aligned radiance fields. All volumetric avatars were computed given only two views as input.}
    \label{fig:novel-views}
\end{figure*}
\begin{figure*}
    \centering
    \includegraphics[trim=0 15 0 5,clip,width=1.0\linewidth]{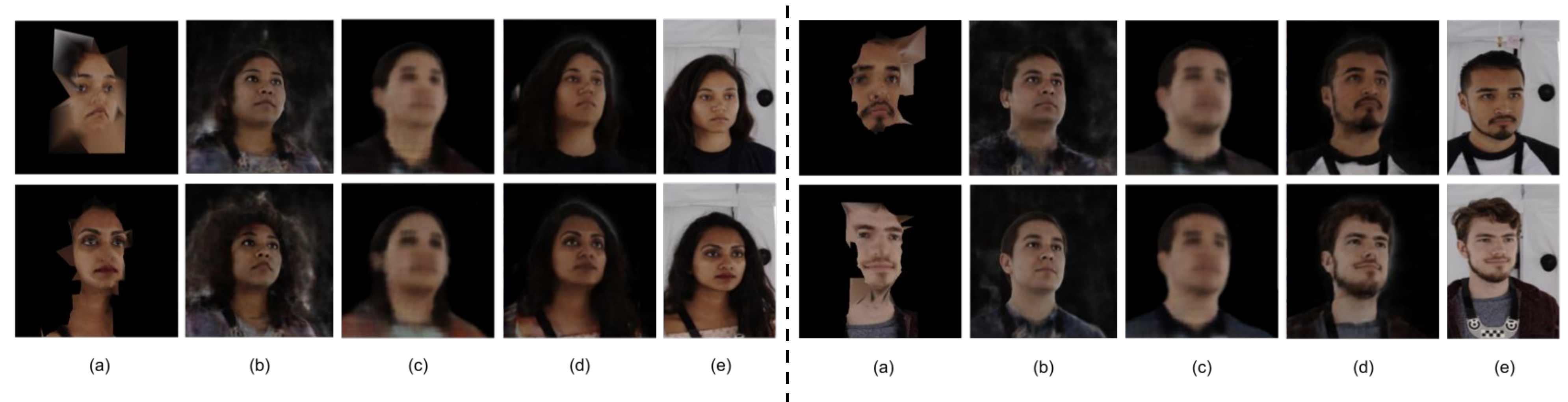}
    \vspace*{-0.5cm}
    \caption{We compare our approach against three baselines: Reality Capture (a), Neural Volumes (b), Globally conditioned NeRF (c). We also show our result (d) and the ground truth identity (e). As can be seen, our approach outperforms the other methods in terms of completeness and level of reconstructed detail by a large margin.}
    \label{fig:comparison}
\end{figure*}
\begin{figure*}
    \centering
    \includegraphics[width=1.0\linewidth]{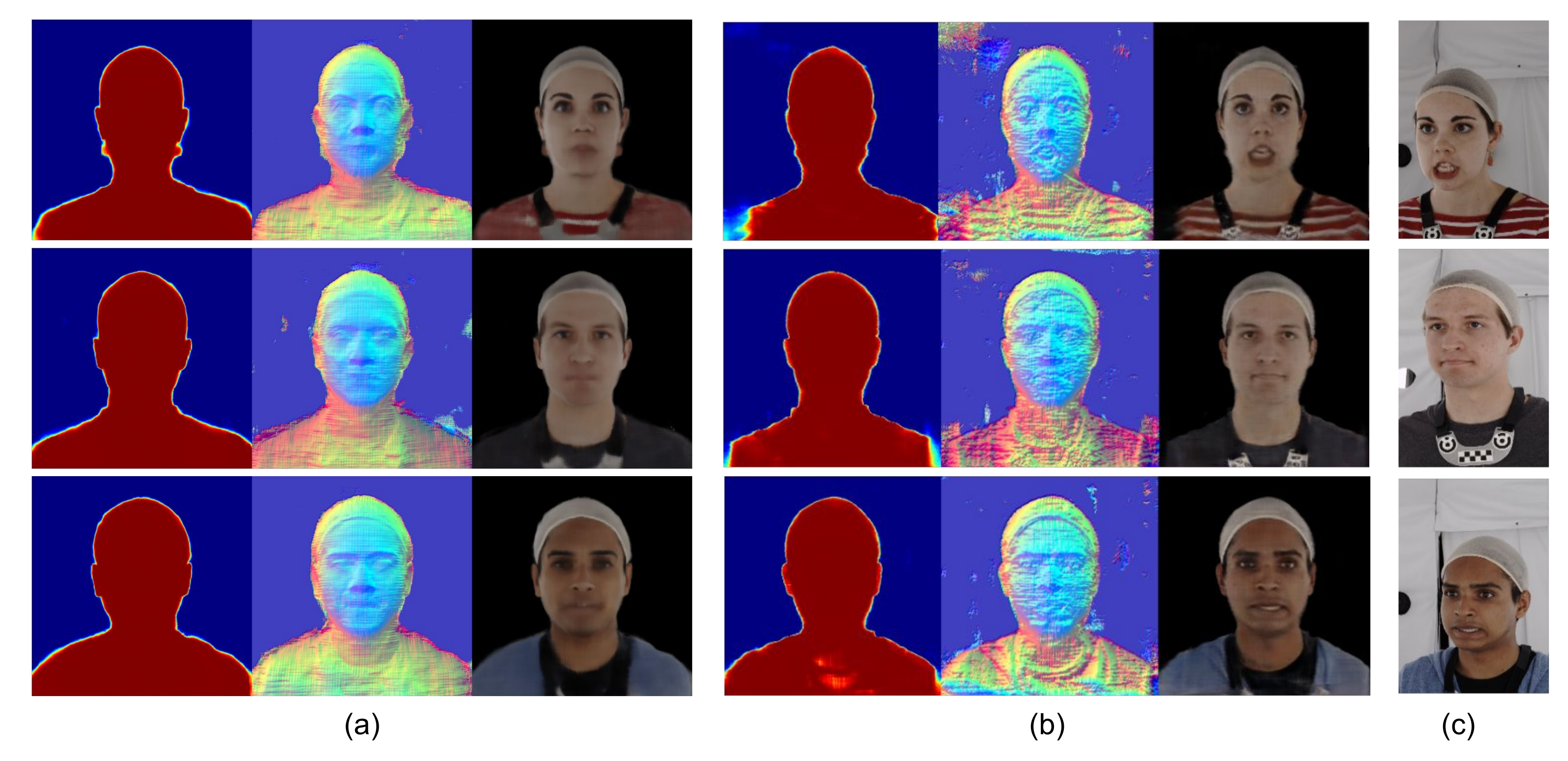}
    \vspace*{-0.8cm}
    \caption{
    Generated alpha/normals/avatar in the canonical viewpoint using (a) eNerf and (b) Ours for (c) the ground truth identity.
    Note, for this experiment the eNeRF baseline has seen all identities and expressions at training time.
    Our approach not only better captures the identity of the person, but also the facial expression, while not having seen these specific identities at training time.
    We attribute this better generalization behaviour to our pixel-aligned features.
    }
    \label{fig:expression}
\end{figure*}
\section{Experiments}
We describe the setup used to capture the training data, describe the baselines used for comparison, and perform quantitative as well as qualitative comparisons.

\subsection{Training Setup}
Our capture setup consists of 53 cameras positioned around the subject. For each subject, we record a set of 30 expressions with a hair-cap. And a neutral expression with no hair-cap. Each frame is fit with a 3D face model including rigid head pose which we use to center the volume between different identities and expressions. We do not use any of the mesh information during training. 
We train our network on 50 subjects using 40 viewpoints and test on held out viewpoints. Additionally, for the expression-based model we train our network on 25 expressions and test on the remaining expressions.
During training, we divide each target image into a $16 \times 16$ grid, and randomly sample a ray from each grid location for a total of $256$ rays per training image. Further, we sample $n_s=128$ points along the ray while clamping the sample points to lie in a unit volume cube. We train our model with a batch-size of 4. Our model takes around 24 hours to converge on 4 Nvidia Tesla V100s.

\subsection{Baselines}
In the following, we introduce the baselines we employ for the qualitative and quantitative comparisons.

\textbf{Reality Capture:}
Is a commercially available software based on classical structure-from-motion (SFM) and multi-view stereo (MVS), that reconstructs a 3d model from a set of captured images. 

\textbf{Neural Volumes:}
Neural volumes (NV) is a voxel-based inference method that globally encodes dynamic images of a scene and decodes a voxel grid and a warp field that represents the scene. 

\textbf{cNeRF:}
We trained a variant of NeRF with global identity conditioning (cNeRF).
Particularly, we employ a VGG-network to extract a single 64D feature vector for each training identity and condition NeRF additionally on this input. 

\subsection{Qualitative Comparisons}
We demonstrate novel view synthesis of unseen identities using our pixel aligned radiance fields, see Fig.~\ref{fig:novel-views}.
As can be seen, given only two views as input, our approach predicts volumetric avatars that can be viewed from a large number of novel viewpoints.

We also compare our method against three baselines that can handle \emph{unseen identities} and \emph{do not use explicit 3D supervision} for training in Fig.~\ref{fig:comparison}.
In all baselines and for our approach, we employ only two images of the novel identities as input to compute the reconstruction.
As can be seen, our approach outperforms all baselines in terms of completeness and the amount of reconstructed details.
Our method produces more complete reconstructions than Reality Capture, which would require many more views of the person to obtain a good reconstruction.
In addition, our approach also leads to more detailed reconstruction than the globally conditioned Neural Volumes and cNerf approaches.
We attribute this better generalization to the use of the pixel-aligned features, that better inform the model at test time.

\subsection{Quantitative Comparisons}
We compare the performance of our method with NV and cNeRF baselines (we omit RC because it fails to capture the complete head shape) in Table~\ref{table:quants} on three common metrics from the literature (SSIM, LPIPS\cite{zhang2018unreasonable} and MSE).
We note that our framework outperforms all the baselines by a considerable margin.
%

\begin{table}[]
    \centering
    \begin{tabular}{c|c|c|c}
         & SSIM ($\uparrow$) &  MSE($\downarrow$) & LPIPS ($\downarrow$) \\
         \hline
        cNeRF & 0.7663 & 1611.0112 & 4.3775 \\
        NV & 0.8027 &  1208.36 & 3.1112 \\
        Ours & \textbf{0.8889} & \textbf{383.71} & \textbf{1.7392} \\
        \hline
    \end{tabular}
    \caption{Quantitative comparison of our approach (Ours) to reconstructions from, Neural Volumes (NV), and Globally conditioned NeRF (cNeRF).}
    \label{table:quants}
\end{table}

\subsection{Analysis}
We observe that methods that use global identity encoding like Neural Volumes and cNeRF do not generalize well to unseen identities as these methods are designed to be trained in a scene specific manner.
Particularly, we notice in cNeRF that the facial features are smoothed out and some of the local details of unseen identities (like facial hair in row 3 and 4, and hair length in row 2) are lost, since this model relies heavily on the learned global prior.
Reality Capture fails to capture the structure of the head as there are no priors built into the SfM+MVS framework, leading to incomplete reconstructions.
A large number of images would be required to faithfully reconstruct a novel identity using RC (we refer to the supplementary document for additional analysis).
Neural volumes is able to generate better textures because of the generated warp field which accounts for some degree of local information. However, since neural volume uses an encoder-decoder architecture, with the encoder using a global encoding, it projects test time identities into the nearest training time identity leading to inaccurate avatar predictions.
Our proposed framework is able to reconstruct volumetric heads from just two example viewpoints, along with the structure of the hair. \\

\textbf{Expression Information}
We present additional qualitative comparison on the ability of our model to better capture expression information in Fig.~\ref{fig:expression}.
We train another conditional NeRF baseline for expressions.
Particularly, since cNeRF cannot generalize to novel identities, we train a NeRF model conditioned on a one-hot expression code and one-hot identity information (eNeRF) on test time identities (unseen for our method).
We observe in this case that despite having seen all the identities during training eNeRF fails to generalize to dynamic expressions for multiple identities.
Since our method leverages the local features for conditioning, it is better able to capture dynamic effects on a specific identity (both geometry and texture).

\begin{figure}[h!]
    \centering
    \includegraphics[width=1.0\linewidth]{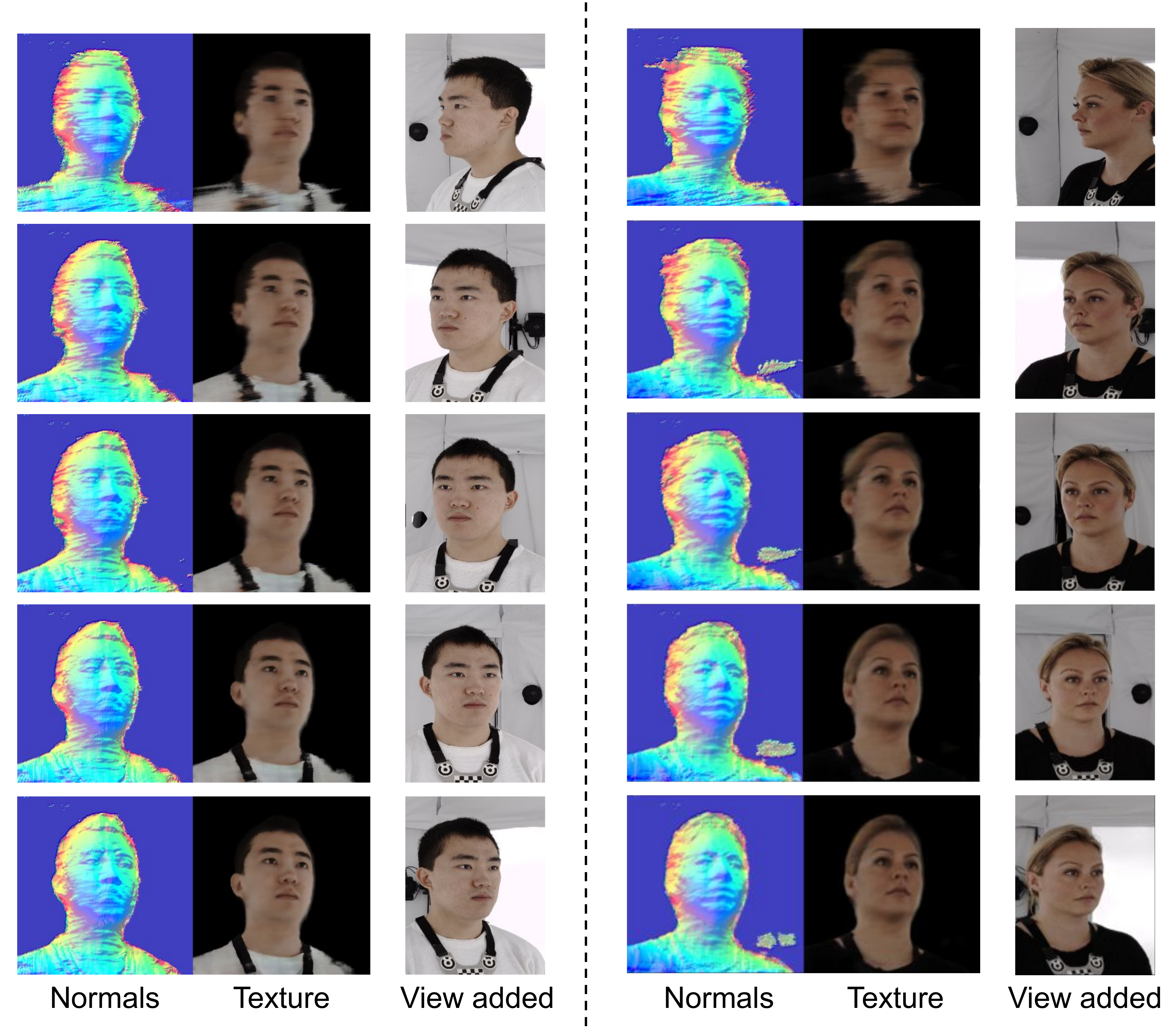}
    \vspace*{-0.3cm}
    \caption{Predicted texture with respect to the number of views. In each row, we add one additional conditioning view (top to bottom). As can be seen, each added input increases the reconstruction quality.}
    \label{fig:num-views}
    \vspace*{-0.2cm}
\end{figure}
\section{Ablation Studies}
In the following, we perform several ablation studies to explore different aspects of our approach in more detail.

\noindent\textbf{How does the quality of the generated images change with respect to the number of example images?}
Fig.~\ref{fig:num-views} shows view extrapolation for unseen identities.
Particularly, since our model learns shape priors from training identities, the predicted normals are consistent with the input identity.
However, when extrapolating to extreme views (1st row), artifacts appear in the parts of the face that are unseen in the conditioning images.
This is because of the inherent depth ambiguity due to projection of the sample points onto the feature image.
We see that adding just the second view already significantly reduces these artifacts as the model now has more information regarding features from different views and can thus reason about depth.
In practice, we find that we can achieve a large degree of view extrapolation with just two conditioning views.

\begin{figure}
    \centering
    \includegraphics[width=1.0\linewidth]{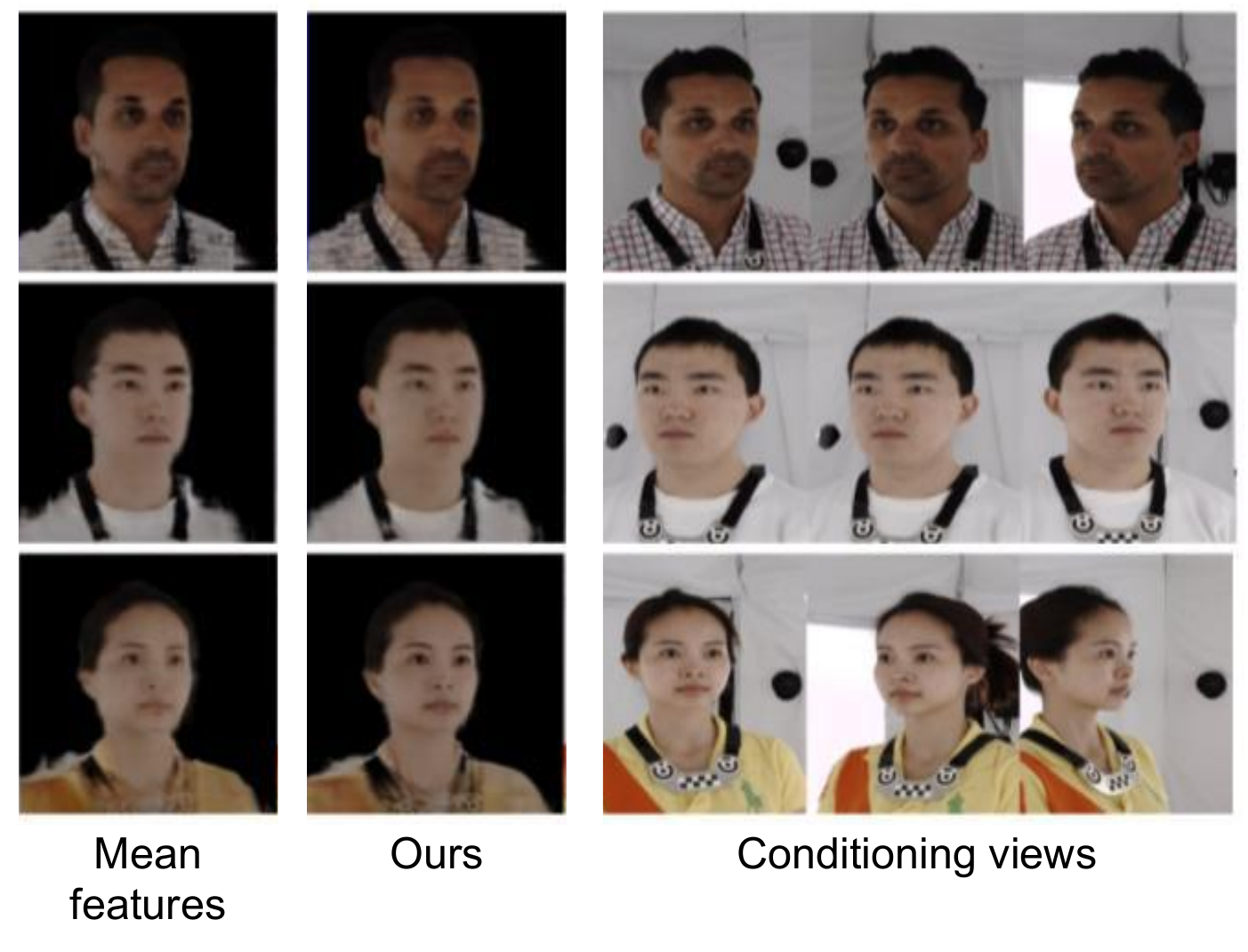}
    \vspace*{-0.2cm}
    \caption{Feature summarization. As can be seen, our camera-aware feature summarization strategy leads to higher quality results than using simple mean pooling.}
    \label{fig:camera-sum}
    \vspace*{-0.2cm}
\end{figure}

\noindent\textbf{Is camera information required in addition to the extracted features?}
Fig.~\ref{fig:camera-sum} demonstrates the need to incorporate camera information in the extracted features.
Particularly, without the camera information, we see a large degree of streaking in the generated images due to inconsistent averaging of information from different viewpoints (particularly in row 1 and 2). 
\begin{figure}
    \centering
    \includegraphics[width=0.95\linewidth]{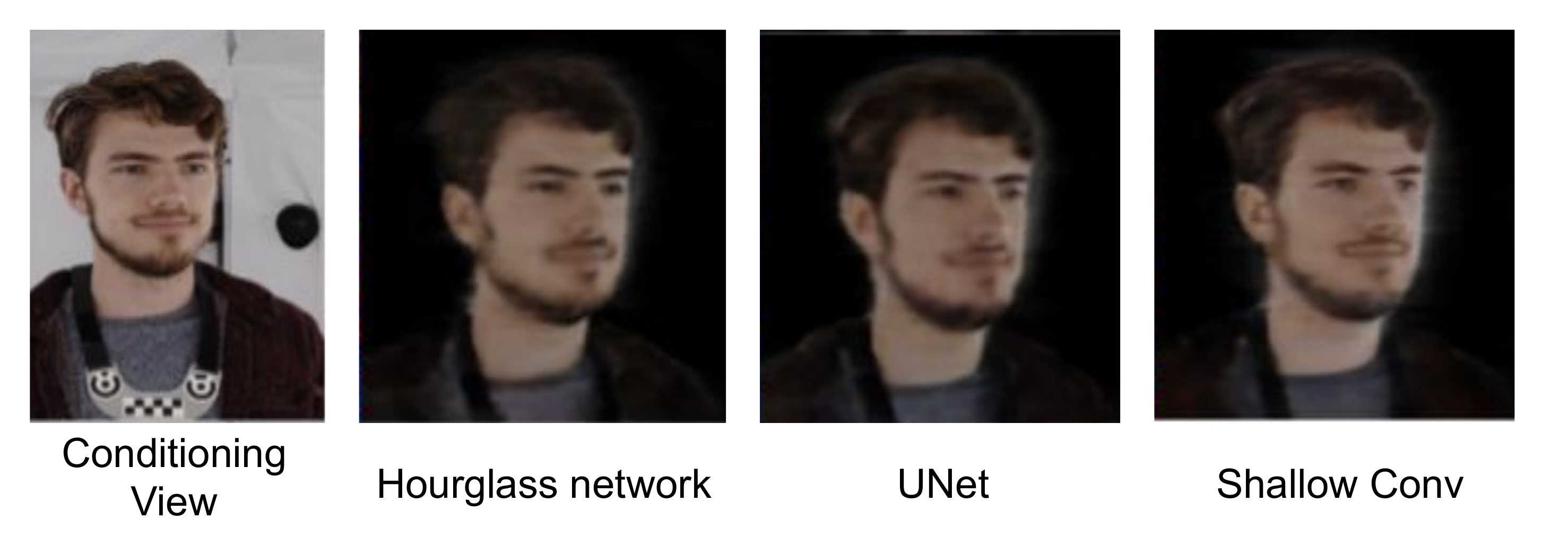}
    \vspace*{-0.4cm}
    \caption{We demonstrate the sensitivity of the pixel-aligned features to the choice of the employed feature extractor. As can be seen, our shallow convolutional network leads to better reconstructions.}
    \label{fig:sensitivity}
\end{figure}

\noindent\textbf{Are the results sensitive to the employed feature extraction network?}
U-Net and hour glass networks are some of the popular feature extraction networks used in recent works \cite{Saito2020}.
However, we find that in our setting a shallow encoder-decoder architecture serves as the best feature extraction networks (Fig.~\ref{fig:sensitivity}) as it preserves more of the local information without having to encode all the pixel level information into a bottleneck layer.







\section{Limitations}
While we have demonstrated compelling results for predicting volumetric avatars of human heads from just a small number of example images, our approach is still subject to a few limitations that can be addressed in follow-up work:
(1) Our approach currently has limited extrapolation capabilities in terms of completely unobserved regions, e.g., the back of the head will not be reconstructed in detail if only front views are provided as example images.
The incorporation of a global prior could improve generalization in such scenarios.
(2) Our approach can currently not be applied to in-the-wild data.
This has multiple reasons:
First, we require the absolute head pose at test time for each of the example images.
Second, our training corpus does not capture the spectrum of illumination and background variation of in-the-wild images.
This could be tackled in the future by a more sophisticated training corpus or by data augmentation strategies.

\section{Conclusion}
We presented PVA - a novel approach for predicting volumetric avatars of the human head given only a small number of images as input.
To this end, we devised a neural radiance field that leverages local, pixel-aligned features that can be extracted directly from the inputs, thus side-stepping the need for very deep or complex neural networks.
Our approach is trained in an end-to-end manner solely based on a photometric re-rendering loss \emph{without requiring} explicit 3D supervision.
We have demonstrated that our approach outperforms the existing state of the art in terms of quality and that we are able to generate faithful facial expression in a multi-identity setting.
We hope that this approach will serve as a simple and strong baseline for future work.

{\small
\bibliographystyle{ieee_fullname}
\bibliography{egbib}
}

\end{document}